\documentclass[runningheads]{llncs}
\usepackage[T1]{fontenc}
\usepackage{graphicx}
\usepackage{booktabs}

\usepackage{microtype}
\usepackage{amsfonts,amsmath, amssymb}
\usepackage{booktabs} 
\usepackage{subfigure}
\usepackage{capt-of}
\usepackage{graphicx}

\usepackage{tocbibind} 
\usepackage[toc,page]{appendix} 
\usepackage{afterpage}
\newcommand\blfootnote[1]{%
  \begingroup
  \renewcommand\thefootnote{}\footnote{#1}%
  \addtocounter{footnote}{-1}%
  \endgroup
}

\usepackage[absolute]{textpos}
\begin{document}

\begin{textblock}{12}(5,1)
\noindent\small 2022 Explainable and Transparent AI and Multi-Agent Systems (EXTRAAMAS 2022)
\end{textblock}

\title{Evaluation of importance estimators in deep learning classifiers for Computed Tomography}
\titlerunning{Importance estimators in deep learning for CT}
%
\author{Lennart Brocki${^a }$\inst{1} \and
Wistan Marchadour${^a }$\inst{2,3} \and
Jonas Maison \inst{2,4} \and
Bogdan Badic \inst{2}
Panagiotis Papadimitroulas\inst{5} \and
Mathieu Hatt ${^b }$\inst{2,*} \and
Franck Vermet ${^b }$\inst{3,*} \and
Neo Christopher Chung ${^b }$\inst{1,*}
}
\authorrunning{Brocki et al.}
%
\institute{Institute of Informatics, University of Warsaw, Warsaw, Poland \and
LaTIM, INSERM, UMR 1101, Univ Brest, Brest, France \and
LMBA, CNRS, UMR 6205, Univ Brest, Brest, France \and
Aquilab, Lille, France \and 
Bioemission Technology Solutions - BIOEMTECH, Athens, Greece\blfootnote{* Corresponding authors: \email{hatt@univ-brest.fr}, \email{Franck.Vermet@univ-brest.fr}, \email{nchchung@gmail.com}}
\blfootnote{a, b : these authors contributed equally
}
}

\maketitle              
\begin{abstract}
Deep learning has shown superb performance in detecting objects and classifying images, ensuring a great promise for analyzing medical imaging. Translating the success of deep learning to medical imaging, in which doctors need to understand the underlying process, requires the capability to interpret and explain the prediction of neural networks. Interpretability of deep neural networks often relies on estimating the importance of input features (e.g., pixels) with respect to the outcome (e.g., class probability). However, a number of importance estimators (also known as saliency maps) have been developed and it is unclear which ones are more relevant for medical imaging applications. In the present work, we investigated the performance of several importance estimators in explaining the classification of computed tomography (CT) images by a convolutional deep network, using three distinct evaluation metrics. Specifically, the ResNet-50 was trained to classify CT scans of lungs acquired with and without contrast agents, in which clinically relevant anatomical areas were manually determined by experts as segmentation masks in the images. Three evaluation metrics were used to quantify different aspects of interpretability. First, the model-centric fidelity measures a decrease in the model accuracy when certain inputs are perturbed. Second, concordance between importance scores and the expert-defined segmentation masks is measured on a pixel level by a receiver operating characteristic (ROC) curves. Third, we measure a region-wise overlap between a XRAI-based map and the segmentation mask by Dice Similarity Coefficients (DSC). Overall, two versions of SmoothGrad topped the fidelity and ROC rankings, whereas both Integrated Gradients and SmoothGrad excelled in DSC evaluation. Interestingly, there was a critical discrepancy between model-centric (fidelity) and human-centric (ROC and DSC) evaluation. Expert expectation and intuition embedded in segmentation maps does not necessarily align with how the model arrived at its prediction. Understanding this difference in interpretability would help harnessing the power of deep learning in medicine.

\end{abstract}

\section{Introduction}
Deep learning models have shown high performance in a plethora of computer vision tasks in the last decade. This impressive performance, however, comes at the cost of lacking transparency, explainability and interpretability of their decision making process \cite{samek2019explainable}. This is particularly problematic in safety-sensitive fields such as health care \cite{papadimitroulas2021artificial,stiglic2020interpretability}, as due to the end-to-end architecture employed, a model's behavior is usually not fully testable and it can fail in unexpected ways \cite{doshi2017towards}. In medical image analysis, it is therefore beneficial to consider interpretability methods that can quantify the relative importance of input features with respect to the class probability, and thus allow to obtain a better understanding of the model's decisions making process. In this work, we aimed at evaluating several such importance estimators in a simple application, namely classifying computed tomography (CT) images acquired with or without contrast agent.

In medical imaging, a contrast agent is a substance ingested by the patient or injected in order to generate contrast enhancement and visibility increase of anatomical structures or blood vessels \cite{lohrke201625}. In our chosen application of CT images of the lungs, an iodine-based solution is usually used and contrast-enhanced images can be quite easily visually identified based on the resulting higher contrast between blood vessels and surrounding tissues \cite{bae2010intravenous,dong2021heavy}. The extraction and exploitation of quantitative metrics from medical images for the purpose of building diagnostic and predictive models is called radiomics \cite{hatt_radiomics_2019}. Radiomic features (shape descriptors, intensity metrics or textural features) have been shown to provide relevant information for diagnostic and prognostic tasks in various applications \cite{hatt_radiomics_2019}. Over the last few years, deep learning techniques have become an important component of radiomics developments, including with end-to-end architectures \cite{diamant_deep_2019}, however the resulting models are notoriously less interpretable than models based on a combination of usual radiomic features \cite{papadimitroulas2021artificial}. 

In the present work, the contrast agent detection/classification task was performed by a trained ResNet-50 architecture modified to accept gray scale CT scan slices as input, while the output layer consists of only one neuron, coding for both without or with contrast agent decisions, with a value respectively closer to 0 or 1. The dataset used for training and testing is a combination of the Lung Image Database Consortium image collection (LIDC-IDRI) \cite{armato_lung_2011} and A Lung Nodule Database (LNDb) \cite{pedrosa_lndb_2019}. In total, 1312 patients scans are available, with more than 320k slices. The model was trained using 2074 slices obtained by selecting 2 slices in each of the 1037 patient 3D scans. On a test set consisting of 4144 slices from 259 different patients, the model obtained 99.4\% accuracy. We emphasize on the fact that the task was chosen to be easy and our focus was obviously not on the training of the model itself. Instead, the task was chosen because experts could quite easily visually identify the parts of the image (i.e., anatomical areas) that they think are relevant to identify whether a contrast agent is present (i.e., parts of the image they first look at when performing the task). This allowed us to explore how well the explanations provided by the importance estimators would match with this human expertise, as is explained in more details below. For the evaluation of importance estimators, 290 slices of the chest, paired to their respective labels, were sampled and the corresponding segmentation masks were produced with the help of a clinical expert.

\begin{figure}
\centering
\includegraphics[width=0.8\textwidth]{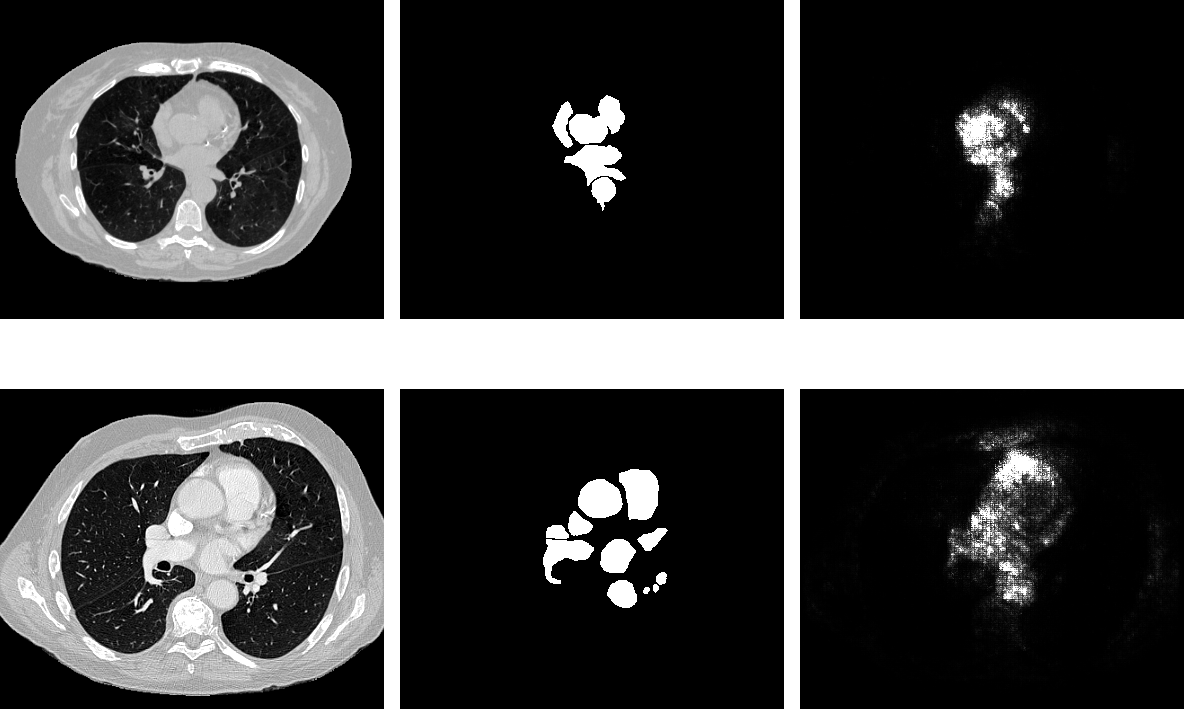}
\caption{\emph{Top row:} Sample of a CT scan slice of chest without contrast agent, the expert-defined segmentation mask and a saliency map obtained using SmoothGrad Squared (from left to right). \emph{Bottom row:} The same graphics as in the top row, here for a sample slice with contrast agent. Notice how areas such as the aorta and the heart regions are highlighted in this case.} \label{fig1}
\end{figure}

A popular approach to mitigate the challenge of interpretability in deep learning is to estimate the importance of input features (i.e., pixels) with respect to the model's predictions. In the literature a wide variety of such importance estimators, also referred to as saliency maps, has been proposed, see for instance \cite{simonyan2014deep,smilkov2017smoothgrad,sundararajan2017axiomatic,zeiler2014visualizing}. The introduction of new saliency map methods is often motivated by certain properties such as sparsity and absence of noise \cite{kim2019saliency,smilkov2017smoothgrad}. That such properties are seen as desirable is mainly motivated by our human expectation of what the model \textit{should} be doing, whereas interpretability methods are supposed to explain what the model \textit{is} doing. An underlying issue is that it is usually not possible to directly evaluate which of these approaches explains a given model most accurately because there is no ground truth available for the model behavior. If there was, it would not be necessary to estimate the importance of input features in the first place. Although a real ground truth is not achievable, a \emph{proxy} gold standard was used here in the form of human-annotated masks (Figure \ref{fig1}). These masks encode the expectation of what regions of the image the network \textit{should} use to predict the presence or absence of contrast agent, at least if we assume that it should rely on the same information as the human expert.

To evaluate importance estimators despite this lack of a ground truth, we considered three different metrics. The design of our computational experiments allowed us to explore and compare model-sensitive evaluation metrics and ones that focus on human expectation and intuition in the context of a medical application. First, in the perturbation-based evaluation, according to a given set of importance scores, we masked input pixels deemed the most important first (MiF) or the least important first (LiF), and measure the resulting accuracy decrease of the model \cite{petsiuk2018rise,samek2016evaluating}. The area between MiF and LiF curves was used as an evaluation metric called \emph{fidelity} \cite{brocki2022_fidelity} (figure \ref{fig2}). Second, segmentation masks were compared against given importance scores on a pixel level. Receiver operating characteristic (ROC) curves and the area under the curve (AUC) approximated how well the importance estimators matched human expectation about the importance of input pixels. Third, importance scores images were segmented to obtain region-wise maps based on the XRAI methodological framework \cite{kapishnikov2019xrai} that can be partially occluded. Dice similarity coefficients (DSC)\cite{dice1945} were computed to assess the overlap of those segmented regions with the reference masks.


\section{Importance estimators}
To interpret classification of deep neural networks for CT scans, we employed a family of importance estimators, also known as saliency maps. We focused on pixel-level methods that quantify importance of input pixels with respect to output probabilities. The majority of applied methods are based on classical gradients (e.g., vanilla saliency maps) \cite{baehrens2010explain,simonyan2014deep}, except Deconvolution \cite{zeiler2014visualizing,zeiler2010deconvolutional}. In the next section we briefly describe the importance estimators under evaluation.

All estimators produce negative and positive importance scores, except for Squared SmoothGrad which produces only positive ones. Many studies routinely use the absolute values of importance scores in their applications, e.g. \cite{simonyan2014deep}. It is however \textit{a priori} unclear to what degree the sign of importance scores is meaningful and contributes to explain the model. When evaluating the different importance estimators we therefore considered not only the original scores but also their absolute values. The following estimators were evaluated:
\begin{itemize}
  \item \textbf{Backpropagation} \cite{baehrens2010explain,simonyan2014deep}: Gradients of the class score $S_c$\footnote{The class score $S_c$ is the activation of the neuron in the prediction vector that corresponds to the class $c$} with respect to input pixels $x_i$
  \begin{align*}
    \mathbf e = \frac{\partial S_c}{\partial x_i}
  \end{align*}
  \item \textbf{Deconvolution} \cite{zeiler2014visualizing,zeiler2010deconvolutional}: Implementation of a mirror version of the trained network, where:
    \begin{itemize}
        \item the optimized weights are transferred,
        \item transposed convolutions are applied,
        \item activation functions are applied on deconvoluted data,
        \item pooled matrices are unpooled using maximum locations memorization.
    \end{itemize}
  \item \textbf{Integrated Gradients} \cite{sundararajan2017axiomatic}: Average over gradients obtained from inputs interpolated between a reference input $x'$ and $x$
  \begin{equation*}
\mathbf{e}=\left({x}_{i}-{x}_{i}^{'}\right) \times \sum_{k=1}^{m} \frac{\partial S_c\left({x}^{'}+\frac{k}{m}\left({x}-{x}^{'}\right)\right)}{\partial x_i} \times \frac{1}{m},
\end{equation*}
    where $x'$ is chosen to be a black image and $m=25$.
  \item \textbf{Integrated Gradients (Black and White)} \cite{kapishnikov2019xrai}: Variant of the original Integrated Gradients method, using both black and white images as reference. This modification theoretically enables importance of dark pixels, when multiplying with the input image.
  \item \textbf{Expected Gradients} \cite{erion2021improving}: Based on Integrated Gradients, the black reference is replaced with a set of training images, and the interpolation coefficients are randomly chosen
  \begin{equation*}
\mathbf{e}=\underset {{x}^{'} \sim D, \alpha \sim U(0, 1)}{\mathbb{E}}\left[({x}_{i}-{x}_{i}^{'}) \times \frac{\partial S_c({x}^{'}+\alpha \times ({x}-{x}^{'}))}{\partial x_i}\right]
\end{equation*}
  \item \textbf{SmoothGrad} \cite{smilkov2017smoothgrad}: Average over gradients obtained from inputs with added noise
    \begin{equation*}
    \mathbf e =\frac{1}{n} \sum_{1}^{n} \hat{\mathbf e}\left(x+\mathcal{N}\left(0, \sigma^{2}\right)\right),
    \end{equation*}
    where $\mathcal{N}\left(0, \sigma^{2}\right)$ represents Gaussian noise with standard deviation $\sigma$, $\hat{\mathbf e}$ is obtained using Backpropagation and $n=15$.
  \item \textbf{Squared SmoothGrad} \cite{hooker2019benchmark}: Variant of SmoothGrad that squares $\hat{\mathbf e}$ before averaging
    \begin{equation*}
    \mathbf e =\frac{1}{n} \sum_{1}^{n} \hat{\mathbf e}\left(x+\mathcal{N}\left(0, \sigma^{2}\right)\right)^2.
    \end{equation*}
\end{itemize}

Lastly, we created a random baseline, by drawing independent and identically distributed numbers from the uniform distribution as important scores in the same dimension as the input image. This random baseline was compared against the aforementioned estimators.

Considering the affirmations made in Integrated and Expected Gradients papers \cite{erion2021improving,sundararajan2017axiomatic} about the meaning of importance scores sign, we applied a targeted inversion of signs in the heatmaps of those saliency methods, to always have positive scores as important for the prediction made by the network. This also allows effective comparison with all other methods, without advanced processing.

\section{Evaluation methods}
In this section we describe the different metrics used for evaluating the importance estimators. The first method is model-sensitive and the two others are based on a comparison with the segmentation masks. 

\subsection{Model accuracy per input feature perturbation}\label{perturb_method}
\begin{figure}
\centering
\includegraphics[width=0.8\textwidth]{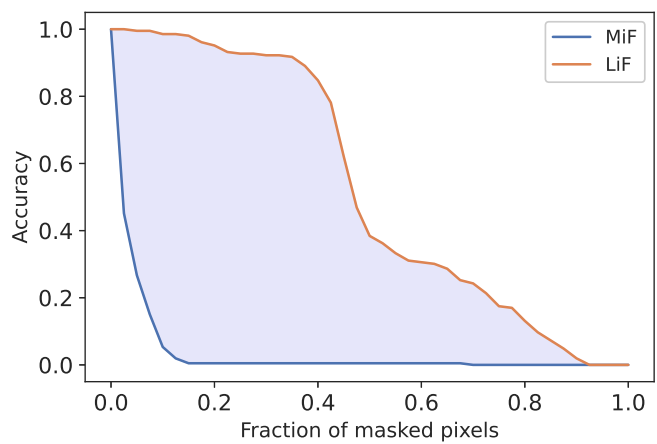}
\caption{The perturbation-based evaluation metric, called \emph{fidelity}, is defined as the area $F$ between the MiF and LiF curves. This does not consider manually annotated segmentation masks in evaluation of importance estimators.} \label{fig2}
\end{figure}

In order to measure how accurately a given importance estimator explains the model we employed a perturbation-based approach \cite{petsiuk2018rise,samek2016evaluating}. To this end the estimated importance scores were first ranked either MiF or LiF and then, according to this ranking, an increasing fraction of pixels, in 2.5 percent steps, was masked. To effectively remove information we chose a label-dependent perturbation, namely for samples with contrast agent pixels were masked with value 0 and in the other case they were masked with value 1. Thereby any information concerning the presence or absence of a contrast agent carried by these pixels was removed. The masked images were then fed to the model and the new accuracy was measured. 

As an evaluation metric, which we called fidelity, we defined the area $F$ between the perturbation curves (see Figure \ref{fig2}), with larger $F$ indicating better performance \cite{brocki2022_fidelity}. An importance estimator is therefore considered to perform well if the model's accuracy drops fast when masking MiF and if simultaneously the accuracy is maintained when masking LiF. With this choice we evaluated if the importance estimators were able to correctly identify both the most important and least important pixels.

\subsection{Concordance between importance scores and segmentation}

Segmentation masks, which indicate the regions relevant for the classification, according to experts, were manually generated. Clinicians may expect that a reasonable classification algorithm for contrast agents would focus on those areas, and therefore these segmentation masks were used to evaluate how well importance estimators match the expert visual process for identifying the presence of contrast agent. We compared the outputs (e.g., importance scores) of interpretability methods against the segmentation masks, to get the receiver operating characteristic (ROC) curve \cite{ESL2011}. More specifically, at an increasing threshold of (normalized) importance scores, we calculated the true positive rate (TPR) and the false positive rate (FPR) for each CT image. Segmentation masks, which were manually annotated by experts, are used such that if a pixel has an importance score above a threshold, it is considered a positive prediction and if that given pixel is within the segmented region, it is declared a true positive. The higher the true positive rate, the more sensitive the estimator is in terms of identifying pixels in the segmented region. We may plot ROC curves for all samples for which masks are provided. We then averaged TPR and FPR at each threshold to obtain the overall ROC for each method. The area under the curve (AUC) was approximated by the mean of the heights of available points. Greater upward deviation from the diagonal line in ROC and higher AUC values indicate better performance.

\subsection{XRAI-based region-wise overlap comparison}
\begin{figure}[h]
\centering
\includegraphics[width=1.0\textwidth]{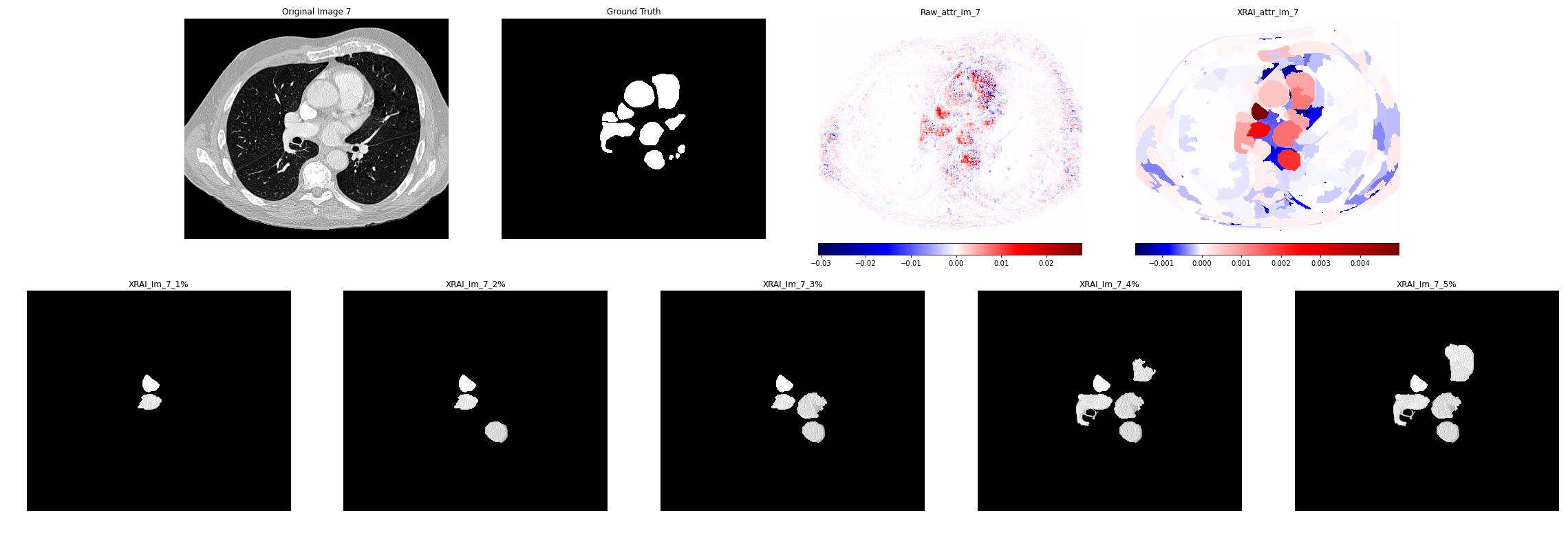}
\caption{\emph{Top row:} The second slice example with contrast agents displayed in Figure \ref{fig1}, its segmentation mask, the heatmap obtained from Expected Gradients, and the segmented version of the heatmap (from left to right). \emph{Bottom row:} Displays of the most salient regions maps (between 1\% and 5\%).} \label{figXRAIorig}
\end{figure}

This metric is based on the XRAI approach \cite{kapishnikov2019xrai}. In its original implementation, the XRAI process is composed of (1) segmentation of the input image using Felzenszwalb’s method \cite{felzenszwalb2004efficient}, (2) computation of Integrated Gradients (Black and White version) importance scores on the input image, (3) ranking of the segmented regions of the image, according to their mean importance scores over the regions, (4) possibility to display any percentage of the most salient regions of the image. An example of this process’ output is shown in Figure \ref{figXRAIorig}.

This algorithm was initially considered as a method for generating saliency maps. We used it here for evaluation purposes by making two observations: (a) any importance estimator can be used instead of the originally proposed Integrated Gradients in the XRAI framework (the ones in our list are compatible); (b) once computed, the XRAI region-wise heatmaps for varying percentages can be directly compared against the available segmentation masks.
From the described modifications, the new process involves variation of the saliency method before segmentation, and partial occlusion of segmented heatmap using a percentage threshold. Partial heatmaps are then evaluated using the Dice similarity coefficient (DSC)\cite{dice1945}, describing the overlapping between two samples (i.e. images in this paper):

\begin{equation*}
DSC=\frac{2 \times |X \cap Y|}{|X| + |Y|}.
\end{equation*}

\section{Results}
\subsection{Model accuracy per input feature perturbation}
\begin{figure}[ht]
\centering
\includegraphics[width=1.0\textwidth]{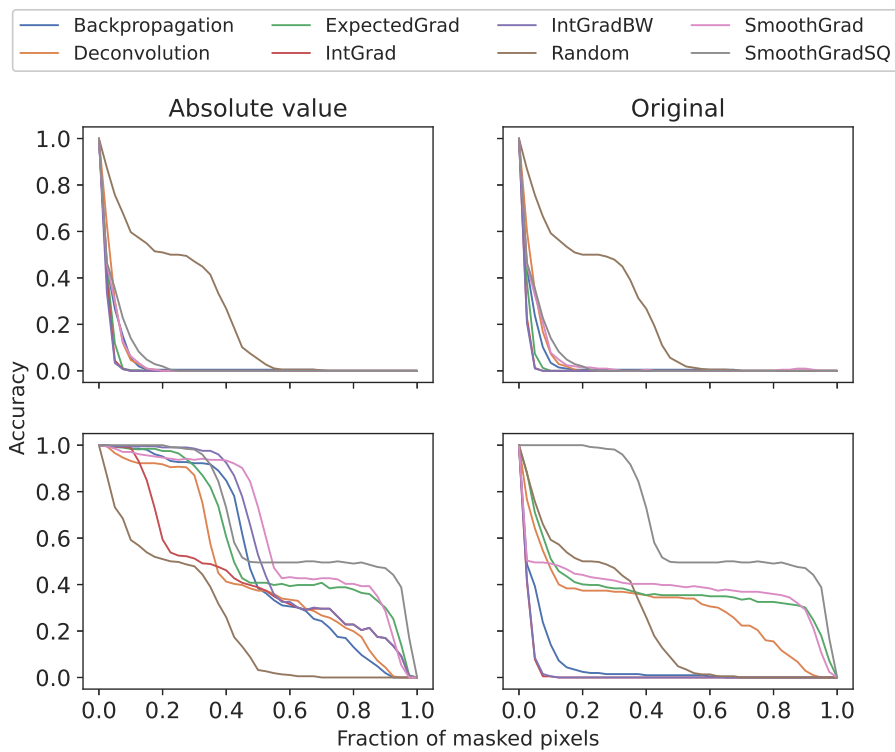}
\caption{Perturbation curves when pixels are masked MiF (\emph{top row}) and LiF (\emph{bottom row}). Notice how for the LiF curves the performance is much better when using absolute values. Overall fidelity metric, as calculated as the area between MiF and LiF curves, are shown for all estimators in Table \ref{fidelitytable}.} \label{fig4}
\end{figure}

\begin{table}[ht]
 \caption{Fidelity metric with and without application of absolute value in the post-processing of importance scores.}
  \label{fidelitytable} 
\vskip 0.15in
\begin{center}
\begin{small}
 \setlength{\tabcolsep}{2em}
 \begin{tabular}{l c c}
 \toprule
 Estimator \hspace{10cm}& $F\times 10$ (Original) & $F\times 10$ (Absolute) \\ [0.5ex] 
 \midrule
 SmoothGradSQ & \textbf{6.2} & \textbf{6.2}\\ 
 SmoothGrad & 3.4 &6.1\\
 IntGradBW & 0.0 & 5.7\\ 
 ExpectedGrad & 3.6 & 5.6\\
 Backprogagation & 0.1 & 4.9\\ 
 Deconvolution & 2.6 & 4.4\\
 IntGrad & 0.0 & 4.2\\
 Random & 0.0 & 0.0 \\ [1ex] 
 
 \bottomrule
 \end{tabular}
 \end{small}
 \end{center}
 \vskip -0.1in
\end{table}

The samples in the test set were masked according to the procedure described in section \ref{perturb_method} and then fed into the model to evaluate its prediction accuracy and obtain the perturbation curves. Since the test set was unbalanced the total model accuracy was obtained from averaging the prediction accuracies for each class. 

The results of this procedure, using fidelity as a metric, are summarized in Table \ref{fidelitytable}. All methods clearly outperformed the random baseline when absolute values were used, with Squared SmoothGrad performing best, closely followed by standard SmoothGrad. Every method led to better results when using the absolute values, with several methods scoring only zero fidelity when the original scores were used.

To gain a better understanding of this behavior we plotted the perturbation curves in Figure \ref{fig4}. Notice how for both versions of Integrated Gradients and Backpropagation the accuracy quickly dropped for LiF masking when the original importance scores were used, just as they did in the MiF case. In fact, all methods, with the exception of Squared SmoothGrad, showed an accelerated  accuracy decrease when the original importance scores were used. This indicates that the pixels with negative importance score are actually evidence contributing to the correct prediction and are not counter-evidence, as one might expect. 

\subsection{Concordance between importance scores and segmentation}

\begin{figure}
\centering
\includegraphics[width=\textwidth]{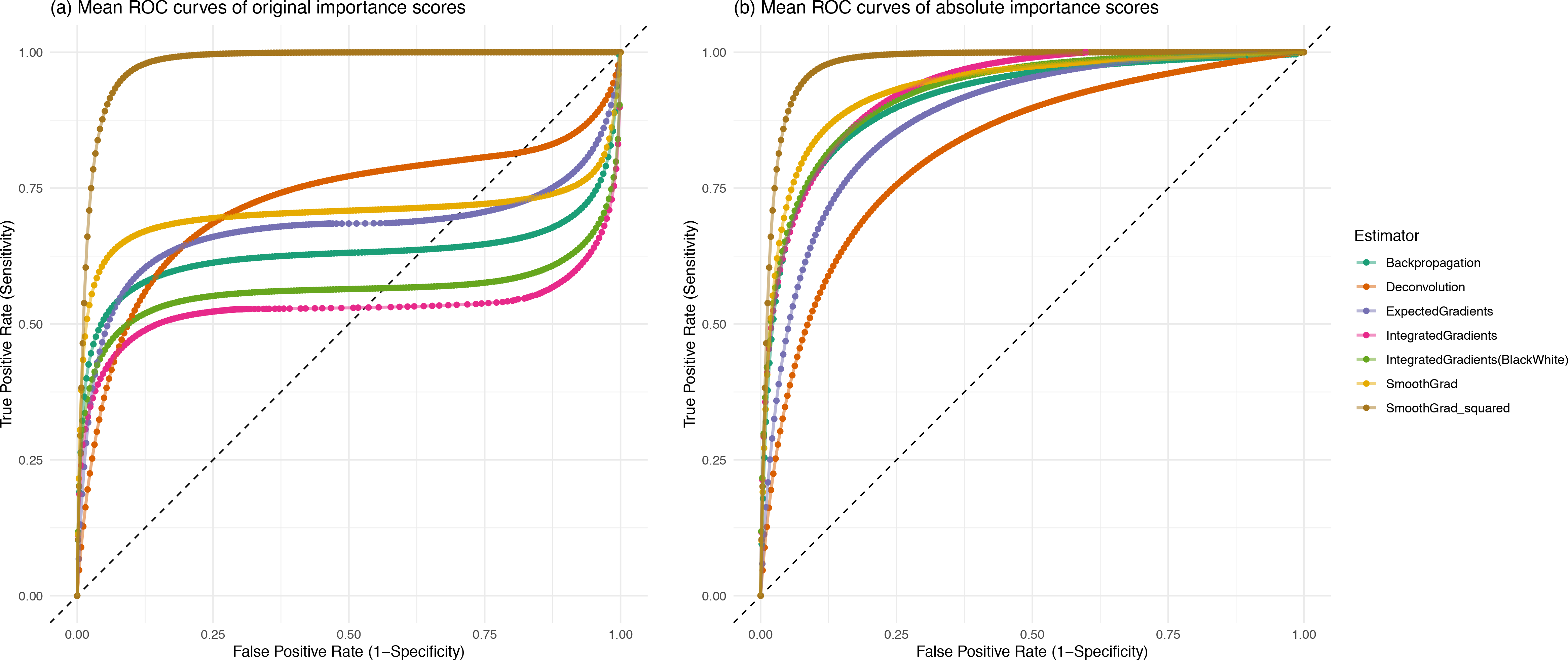}
\caption{Concordance between importance scores and segmentation, measured by receiver operating characteristic (ROC) curves. ROC curves were averaged at increasing thresholds of normalized importance scores. (a) The original importance scores, as outputted by interpretability methods, were used directly to compute ROC curves. (b) The absolute values of importance scores were used.} \label{ROC_mean}
\end{figure}

The segmentation masks were compared against the importance scores. For each CT sample, the true positive rate (TPR) and the false positive rate (FPR) were calculated and plotted against. We used both the original importance scores and the absolute importance scores. We obtained the overall ROC curves by taking means of TPRs and FPRs at different thresholds (Figure \ref{ROC_mean}).

To quantify upward deviation from the random baseline (diagonal line), we estimated the area under the curve (AUC) for each importance estimator. In the absolute scale, Squared SmoothGrad and SmoothGrad performed best, followed by Integrated Gradients (Table \ref{AUCROC}). In the original scale, the AUC from Squared SmoothGrad was the highest, followed by Deconvolution. Note that for almost all considered estimators, relying on the absolute values increased AUCs, although some benefited more than others. The only exception was Squared SmoothGrad, which was the only method that resulted in non-negative original scores and its AUC was therefore identical in both. 

\begin{table}[ht]
 \caption{Area under the curve (AUC) from receiver operating characteristic (ROC) based on concordance between importance scores and segmentation}
  \label{AUCROC} 
\begin{center}
\begin{small}
\setlength{\tabcolsep}{2em}
\begin{tabular}{lrr}
  \toprule
  Estimator & AUC (Original) & AUC (Absolute) \\ 
 \midrule
   SmoothGradSQ & \textbf{0.96} & \textbf{0.96} \\ 
   SmoothGrad & 0.70 & 0.92 \\ 
   IntGrad & 0.53 & 0.91 \\ 
   IntGradBW & 0.56 & 0.91 \\ 
   Backpropagation & 0.62 & 0.90 \\ 
   ExpectedGrad & 0.67 & 0.87 \\ 
   Deconvolution & 0.71 & 0.81 \\ 
    \bottomrule
\end{tabular}
\end{small}
\end{center}
\end{table}

\subsection{XRAI-based region-wise overlap comparison}

\begin{figure}[h]
\centering
\includegraphics[width=1.0\textwidth]{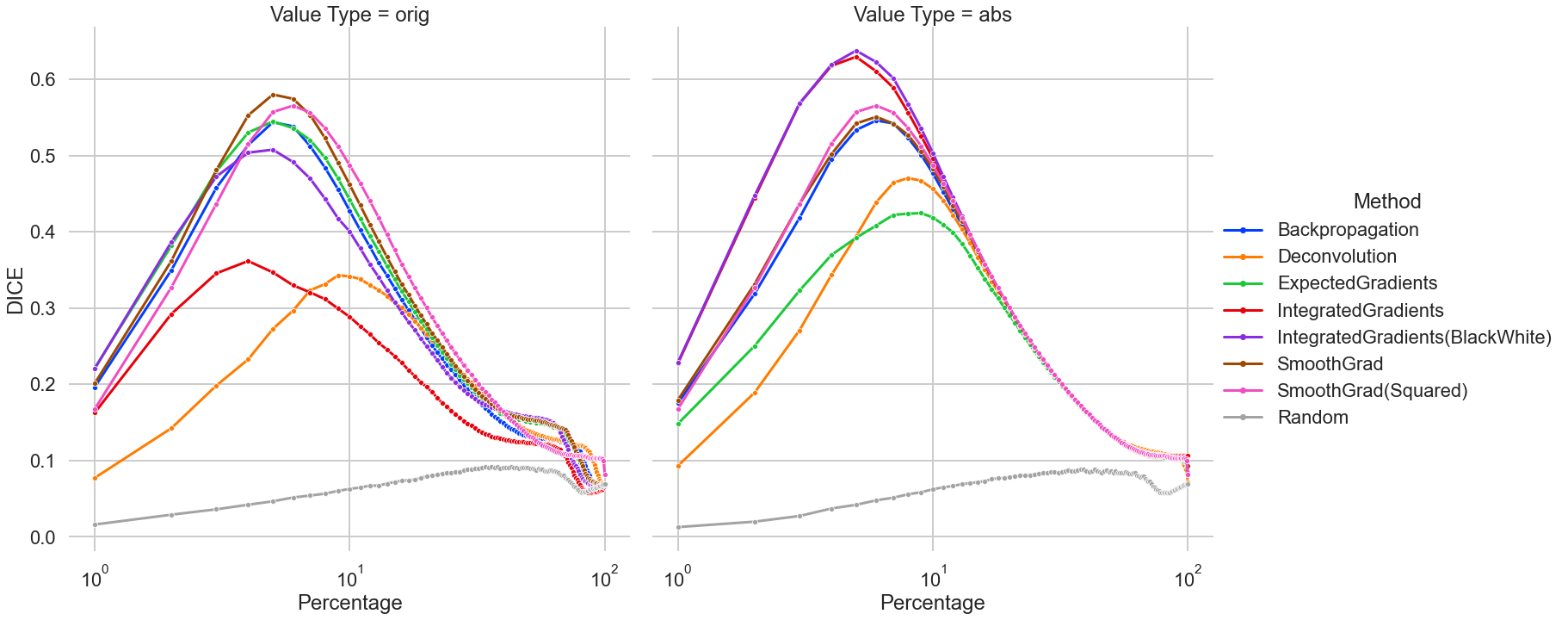}
\caption{Dice Similarity Coefficients against displayed percentage of most salient regions in segmented heatmaps, for original values (left) and absolute values (right). We used log-scale on the x axis, and each data point was the average of coefficients over tested slices.} \label{DICECurvesXRAI}
\end{figure}

The segmented heatmaps obtained from the XRAI algorithm were compared to our \emph{proxy} gold standard masks. The DSC, a similarity metric, is computed from each comparison: an "ideal" heatmap from expectation should display all of the mask sub-regions, while keeping at zero-value the useless regions for network prediction. The averaged coefficients over samples were plotted against the percentage threshold of most salient regions in Figure \ref{DICECurvesXRAI}. The curves for original values and absolute values were compared using the same y axis, and after a first analysis the x axis was configured on a log scale, in order to emphasize on the most informative range of percentages, i.e., from 1\% to 10\%.
In Table \ref{DICEmax} are logged the highest value for every curve in Figure \ref{DICECurvesXRAI}, i.e. when the threshold was optimized to get the best coefficient.

We observed that relying on absolute values led to similar or higher values for all methods, except Expected Gradients, an anomaly analysed in section \ref{discuss}. As expected, the random distribution coefficients were outperformed by all other methods.
When considering the original values, SmoothGrad variants were the more accurate methods, followed closely by Expected Gradients. When relying on absolute values instead, Integrated Gradients variants increased sharply to better performance than SmoothGrad methods and Backpropagation. We also denoted the overall low performance of Deconvolution; considering the heatmaps obtained and the performance loss, it is possible that this method is not fully adapted for segmentation.

The DSC peaks were always achieved between 5\% and 10\% of the threshold, implying that increasing the percentage only adds unimportant regions to the segmented map. This hypothesis is supported by the observed ratio of masked information over the full image size. This value, when averaged over the slices of our dataset, was at 3.6 \% (1-11 \%), meaning that maximum performance could be achieved on this percentage of display.

\begin{table}[h]
 \caption{Maximum values of Dice similarity coefficients from XRAI-based region-wise overlap comparison}
  \label{DICEmax} 
\begin{center}
\begin{small}
\setlength{\tabcolsep}{2em}
\begin{tabular}{lcc}
  \toprule
  Estimator & Max (Original) & Max (Absolute) \\ 
 \midrule
   IntGradBW & 0.51 & \textbf{0.64} \\ 
   IntGrad & 0.36 & 0.63 \\ 
   SmoothGradSQ & 0.57 & 0.57 \\ 
   SmoothGrad & \textbf{0.58} & 0.55 \\ 
   Backpropagation & 0.54 & 0.55 \\ 
   Deconvolution & 0.34 & 0.47 \\ 
   ExpectedGrad & 0.55 & 0.42 \\ 
   Random & 0.09 & 0.09 \\
  \bottomrule
\end{tabular}
\end{small}
\end{center}
\end{table}

\pagebreak
\newpage

\section{Discussion}\label{discuss}
Interpretability of deep neural networks is an immense challenge for their applicability in medical imaging. We are particularly interested in explaining which input features a model utilizes to make its prediction. In this study, a ResNet-50 model was trained on CT scans to classify slices as acquired with or without contrast agents, as a simple task example that could be easily interpreted by human experts, in order to provide a proxy ground-truth for comparison. Selected CT scans were indeed processed by a human expert to define segmentation masks of the relevant anatomical areas most informative of the presence of injected contrast agent. To understand how different importance estimators work, we applied three evaluation methods. While this study suggests certain importance estimators perform better than others for this application domain, it represents an investigation into similarity and differences between model- and human-centric evaluation.

Since almost all of the importance estimators benefited from relying on the absolute values we focus our discussion of their relative performance on this case. Both fidelity (Table \ref{fidelitytable}) and AUC (Table \ref{AUCROC}) agree that SmoothGrad and Squared SmoothGrad performed the best. Using the Dice coefficient (Table \ref{DICEmax}) as evaluation metric, both versions of Integrated Gradients performed best, closely followed by both versions of SmoothGrad. When using black and white references for Integrated Gradients instead of just the black one, only the fidelity score substantially improved. Deconvolution and Expected Gradients overall performed poorly, with the exception that for fidelity Expected Gradients achieved a middle rank. Although the three evaluation metrics roughly agreed on the performance of most importance estimators, there were also important differences. The most glaring one being Integrated Gradients, which performed second best for Dice but worst for fidelity.

As mentioned above, most of importance estimators benefited from using absolute values instead of the original ones. This implies that when the original values are used to rank the pixels, the least important ones actually contain evidence supporting the correct prediction, instead of counter-evidence. This observation may offer an explanation for the presence of saddle points in the ROC curves in Figure \ref{ROC_mean}(a). The only exception is Expected Gradients, whose performance was worse using the absolute values for the XRAI-based evaluation. In the Expected Gradients paper \cite{erion2021improving}, the authors stated that heatmaps show counter-evidence to the network prediction, using the opposite sign. Therefore, keeping the original values should ensure better performance than converting them to absolute values. However, the fact that we observed this only in the XRAI scores has a possible explanation: the use of segmentation on the input image, leading to information loss when averaging over the regions. While truly useful for results clarity and better insight of a network’s behaviour, the XRAI approach creates segmented maps from input image, and not the pixel-wise heatmap. This leads in some cases to a great difference between the segmented areas and visual grouping of importance scores on the heatmap. Some scores are then neutralised by averaging over the regions (information loss), and performance of the process is decreased. This limitation can be addressed using a modified segmentation (on the heatmap instead of input image), or replacing the average over regions by an alternative statistical operation.

The fact that using absolute values clearly has a positive impact on the performance of the importance estimators seems at first glance counter-intuitive. One would expect that negative importance scores indicate counter-evidence and that by using absolute values this information is lost, which would harm their performance. The first thing to notice is that for methods like Backpropagation and SmoothGrad, which are essentially equal to the gradient, one does not expect the sign to be meaningful. The reason for this can be understood by comparison with a simple linear model: the correct attribution is obtained by the product of the input and the gradient(see also discussion in section 3.2 of \cite{ancona2017towards}). Simlarly, for IntGradBW the meaning of the sign is not clear due to the different reference images that are used.
More surprisingly, also IntGrad, which includes a multiplication with the input, clearly improves when absolute values are used. An explanation could be offered by the authors of  \cite{ancona2017towards}: depending on the model, estimators might exhibit low accuracy in predicting the sign of the importance score. Therein it is shown, that for a simple MNIST classifier the sign is predicted with high accuracy, whereas for a more complex model trained on ImageNet this accuracy is low. This appears to be also the case for our model. For each model one should therefore carefully evaluate whether absolute values are beneficial or not.

Our different evaluation metrics reveal how human intuition may differ from model-sensitive evaluation. The fidelity metric does not consider segmentation masks, while the other two evaluation methods do. Segmentation masks were created by manually delineating in CT images the anatomical areas containing a priori the most relevant information, according to clinicians who claim they most easily detect the presence of contrast agent by looking at them. To a certain extent, the models that could be trusted and adapted in medicine may require substantial concordance with expectation of clinicians. On the other hand, deep neural networks may certainly take different routes to arrive at their prediction. It is, for instance, conceivable that the model only uses a small number of input pixels to reach its predictions and that the importance ranking of the remaining pixels is more or less random, in which case one would not expect a good matching of saliency map and segmentation mask. This might explain the differences in the ranking of the estimators we observed between the three evaluation metrics. Furthermore, it is possible that the model utilizes features beyond the segmented regions to make its predictions.

The present work has several limitations for generalization to CT. We investigated a single application, over a single classification task, through a single architecture. The performances of importance estimators under consideration might be different when applied on different tasks or using different network architectures. We are interested in applying our evaluation methods to more general applications in medical imaging. Furthermore, it would be of great interest to obtain more insight into how useful the offered explanations actually are for users. For instance, the region-based explanations such as XRAI potentially lose some accuracy by averaging over the original importance scores but are conceivably easier to interpret for a clinician, compare to the original and segmented heatmaps in figure \ref{figXRAIorig}. We hope to address this question in future work. Nonetheless, our evaluation of diverse importance estimators using distinct metrics for CT represents a critical step towards comprehensive understanding of interpretability in deep learning.

\section*{Acknowledgements}
This work was partly funded by the ERA-Net CHIST-ERA grant [CHIST-ERA-19-XAI-007] long term challenges in ICT project INFORM (ID: 93603), the French Ministry for Research and Higher Education and the French National Research Agency (ANR) through the CIFRE program, The Brittany Region, by General Secretariat for Research and Innovation (GSRI) of Greece, by National Science Centre (NCN) of Poland [2020/02/Y/ST6/00071]. This work has utilized computing resources from the Interdisciplinary Centre for Mathematical and Computational Modelling (ICM) at University of Warsaw and the NVIDIA GPU grant.

\bibliographystyle{splncs04}
\bibliography{extramaas}







\end{document}